\documentclass[10pt,twocolumn,letterpaper]{article}

\usepackage{iccv}
\usepackage{times}
\usepackage{epsfig}
\usepackage{graphicx}
\usepackage{amsmath}
\usepackage{amssymb}
\usepackage{graphicx}
\usepackage{multirow}
\usepackage{booktabs}

\usepackage[pagebackref=true,breaklinks=true,letterpaper=true,colorlinks,bookmarks=false]{hyperref}

\iccvfinalcopy 

\def\iccvPaperID{3902} 

\ificcvfinal\fi

\begin{document}

\title{A Learnable Self-supervised Task for Unsupervised Domain Adaptation on Point Clouds}

\author{Xiaoyuan Luo\thanks{ Equal Contribution.}, Shaolei Liu\footnotemark[1], Kexue Fu, Manning Wang\thanks{Corresponding Authors}, Zhijian Song\footnotemark[2]\\
Digital Medical Research Center, School of Basic Medical Science, Fudan University\\
Shanghai Key Lab of Medical Image Computing and Computer Assisted Intervention\\
{\tt\small \{xyluo19,slliu19,kefu18,mnwang,zjsong\}@fudan.edu.cn}
}

\maketitle
\ificcvfinal\thispagestyle{empty}\fi

\begin{abstract}
   Deep neural networks have achieved promising performance in supervised point cloud applications, but manual annotation is extremely expensive and time-consuming in supervised learning schemes. Unsupervised domain adaptation (UDA) addresses this problem by training a model with only labeled data in the source domain but making the model generalize well in the target domain. Existing studies show that self-supervised learning using both source and target domain data can help improve the adaptability of trained models, but they all rely on hand-crafted designs of the self-supervised tasks. In this paper, we propose a learnable self-supervised task and integrate it into a self-supervision-based point cloud UDA architecture. Specifically, we propose a learnable nonlinear transformation that transforms a part of a point cloud to generate abundant and complicated point clouds while retaining the original semantic information, and the proposed self-supervised task is to reconstruct the original point cloud from the transformed ones. In the UDA architecture, an encoder is shared between the networks for the self-supervised task and the main task of point cloud classification or segmentation, so that the encoder can be trained to extract features suitable for both the source and the target domain data. Experiments on PointDA-10 and PointSegDA datasets show that the proposed method achieves new state-of-the-art performance on both classification and segmentation tasks of point cloud UDA. Code will be made publicly available.
\end{abstract}

\section{Introduction}

Point clouds contain abundant spatial geometric information and have become an indispensable 3D data representation in computer vision. With the continuous advancement of range sensors, such as depth cameras and LIDAR, more and more 3D point clouds are captured and processed in many different applications, including automated driving \cite{64}, human-computer interactions \cite{65} and augmented reality \cite{66}. Recently, deep neural networks have exhibited excellent performance in supervised tasks on point clouds, such as classification \cite{1,2}, segmentation \cite{3,4,5} and registration \cite{6,7,8}. In the supervised learning scheme, manual labeling of a large amount of point clouds is needed for model training. However, manual labeling is often extremely expensive and time-consuming, especially when we need to label point clouds for segmentation.  Therefore, increasing attention has been attracted to techniques that help alleviate the dependence on labeled data.
\begin{figure}[t]
   \centering
   \includegraphics[scale=0.57]{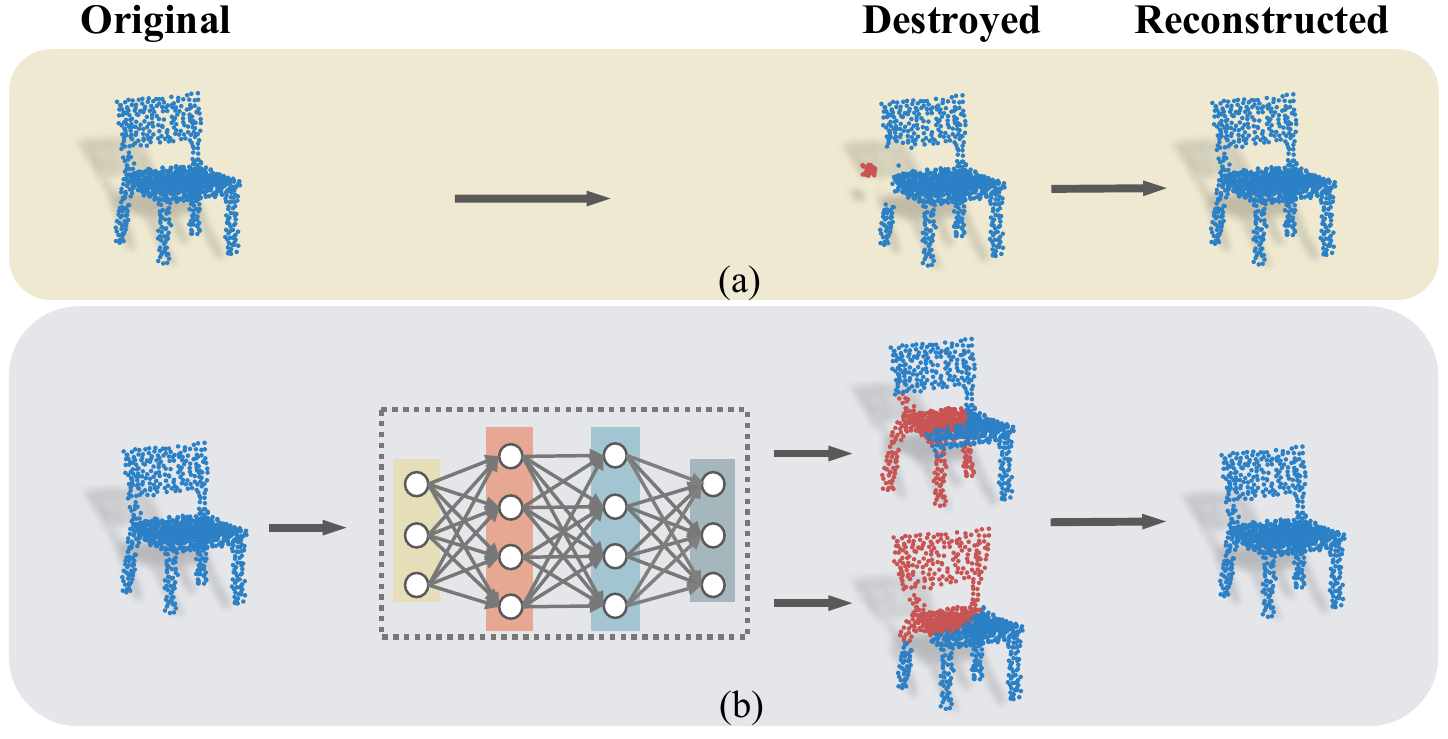}
   \caption{ Destruction-reconstruction self-supervised task on point clouds. The original point cloud is first destroyed by some manner and then a deep neural network is trained to reconstruct the original point cloud from the destroyed ones. The red points indicate the part of the point cloud that is destroyed.  (a) Hand-crafted destruction: replace a part of the point cloud with new points sampled from Gaussian distribution. (b) Proposed learnable destruction: local nonlinear transformation by a deep neural network. }
   \label{figure1}
\end{figure}

Unsupervised domain adaptation (UDA) is one of the methods to address the problem of label-scare dataset, and it aims to leverage labeled source domain data and unlabeled target domain data to train a model that works in the target domain. Because of the inevitable distribution discrepancy between the source domain and the target domain, the model trained only in the source domain usually does not work well in the target domain. In 2D imaging field, many widely used domain adaptation (DA) methods have been proposed \cite{60}, where domain adaptation is usually achieved by globally matching the feature distributions of the source and the target domains \cite{67}. However, in the field of 3D point cloud processing, the feature distributions learned from different domains are quite different \cite{18}, and the DA methods designed for 2D images such as MCD \cite{24} and DANN \cite{68} cannot be transferred directly to point clouds. Recently, a number of studies have emerged for point cloud UDA by learning domain-invariant features. As mentioned in \cite{49},  local features of point clouds are more transferable than global features, so the authors proposed using adaptive nodes to align the local features between the source and the target domains. However, the experimental results on the point cloud domain adaptation benchmark PointDA-10 \cite{49} were not satisfactory. In contrast, \cite{18} and \cite{20} proposed utilizing self-supervised tasks to help capture highly transferable feature representations, and they achieved higher classification accuracy than \cite{49} on PointDA-10, thereby demonstrating the effectiveness of integrating self-supervised tasks in point cloud UDA.

In this paper, we propose a novel learnable self-supervised task and integrate it into a self-supervision-based point cloud UDA architecture \cite{20,76,77} to achieve improved performance. The UDA architecture contains a main network for the main supervised task and an auxiliary network for the auxiliary self-supervised task. The two networks share an encoder but have different heads to perform the two tasks separately. The labeled data in the source domain are used to train the main network, and the data in both the source domain and the target domain are used together to train the auxiliary network, where the pseudo labels used for training are generated automatically. Therefore, the encoder is trained to extract features from data in both the source domain and the target domain so that it can be effectively transferred to the target domain. In this architecture, how well the encoder can be transferred to the target domain heavily depends on the self-supervised task. For this reason, we propose a learnable nonlinear transformation on point clouds and use it to design a destruction-reconstruction self-supervised task,  as shown in Figure \ref{figure1}. Destruction-reconstruction is a commonly used strategy for designing self-supervised tasks, but all existing studies have explored hand-crafted methods to destroy the point cloud, such as the one shown in Figure \ref{figure1} (a) \cite{18}. In contrast, as shown in Figure \ref{figure1} (b), we propose a learnable transformation based on a deep neural network, which destroys a point cloud by applying nonlinear transformation on a part of it. Through adversarial training, the network is able to learn a continuous nonlinear transformation to generate highly abundant and complicated point clouds while retaining the semantic information. By reconstructing the original point cloud from these locally transformed ones, the encoder can learn to extract local features for both the source domain and the target domain data.

Our main contributions are summarized as follows:
\begin{itemize}
   \item[$\bullet$]  We propose a novel learnable transformation on point clouds and construct a learnable point cloud destruction-reconstruction self-supervised task. Compared with existing hand-crafted self-supervised tasks, our method can learn more transferable cross-domain features to mitigate the distribution discrepancy. To our best knowledge, this work is the first learnable self-supervised task for both the point cloud processing and the broader CV field.
   \item[$\bullet$] We apply the proposed learnable self-supervised task in point cloud UDA and developed a multi-region destruction strategy. Destructing point clouds in different regions and then reconstructing the original point clouds encourages the encoder in the UDA architecture to focus on local features, which is beneficial to domain adaptation.
   \item[$\bullet$] The proposed method is evaluated on PointDA-10 and PointSegDA datasets for point cloud classification and segmentation UDA, respectively, and it achieves new state-of-the-art performance on both tasks.
\end{itemize}

\section{Related Woks}

\subsection{Deep Learning on Point Clouds}

With the unprecedented success achieved by deep neural networks in 2D visual tasks, many deep learning techniques have been proposed for 3D point clouds. A major difference between 3D point clouds and 2D images is that point clouds are unordered sets, so the networks for point clouds should be permutation invariant. A series of architectures have been proposed to process 3D point clouds, such as PointNet \cite{1}, PointNet++ \cite{29}, DGCNN \cite{30}, \emph{etc}. Interested readers are referred to a latest survey \cite{77} for details.

\subsection{Domain Adaptation on Point Clouds}

Domain adaptation is a  branch of transfer learning and has long been an important issue in machine learning \cite{62}. In UDA, the source domain and the target domain tasks are the same, but the data distributions of the two domains are different \cite{61}. Meanwhile, we only have labels for the source domain data. The objective of UDA is to train a model that can be used directly on target domain data.

With the rapid development of deep learning techniques in computer vision, deep learning based DA was first developed for 2D images \cite{63} and subsequently for 3D point clouds. Some studies \cite{40,41} applied 2D image DA techniques to 3D point clouds, losing large amounts of 3D geometric information, which is crucial for understanding 3D shapes. In addition, most general-purpose DA methods for 2D images merely perform global feature alignment without utilizing local geometric information, impeding their applications on 3D point clouds. Two types of approaches have been proposed to utilize local information in point cloud DA. One type aligns the local and global features at the same time, and the other type utilizes the powerful feature extraction capability of self-supervised learning to capture more effective local information. PointDAN \cite{49} belongs to the first type, and it jointly aligns the global and local features of  source and target domains at multiple scales, but its performance on PointDA-10 is not very good. In the second type, self-supervised tasks are applied to capture abundant local and global features. The effectiveness of self-supervised learning in promoting DA has been demonstrated in 2D images \cite{76,77}, and it has also been used for point cloud UDA. As the first attempt of this approach, \cite{20} introduces a destruction-reconstruction self-supervised auxiliary task for point cloud UDA, and the destruction is made by replacing several randomly selected regions with new points sampled from an isotropic Gaussian distribution. \cite{18} utilizes similar UDA architecture as \cite{20} but proposes a new self-supervised 3D puzzle auxiliary task for UDA. The above studies show the effectiveness of integrating self-supervised task in point cloud UDA.  In this work, we follow this line of work and propose a new self-supervised task for better domain adaptation.

\begin{figure*}[t]
   \centering
   \includegraphics[scale=0.61]{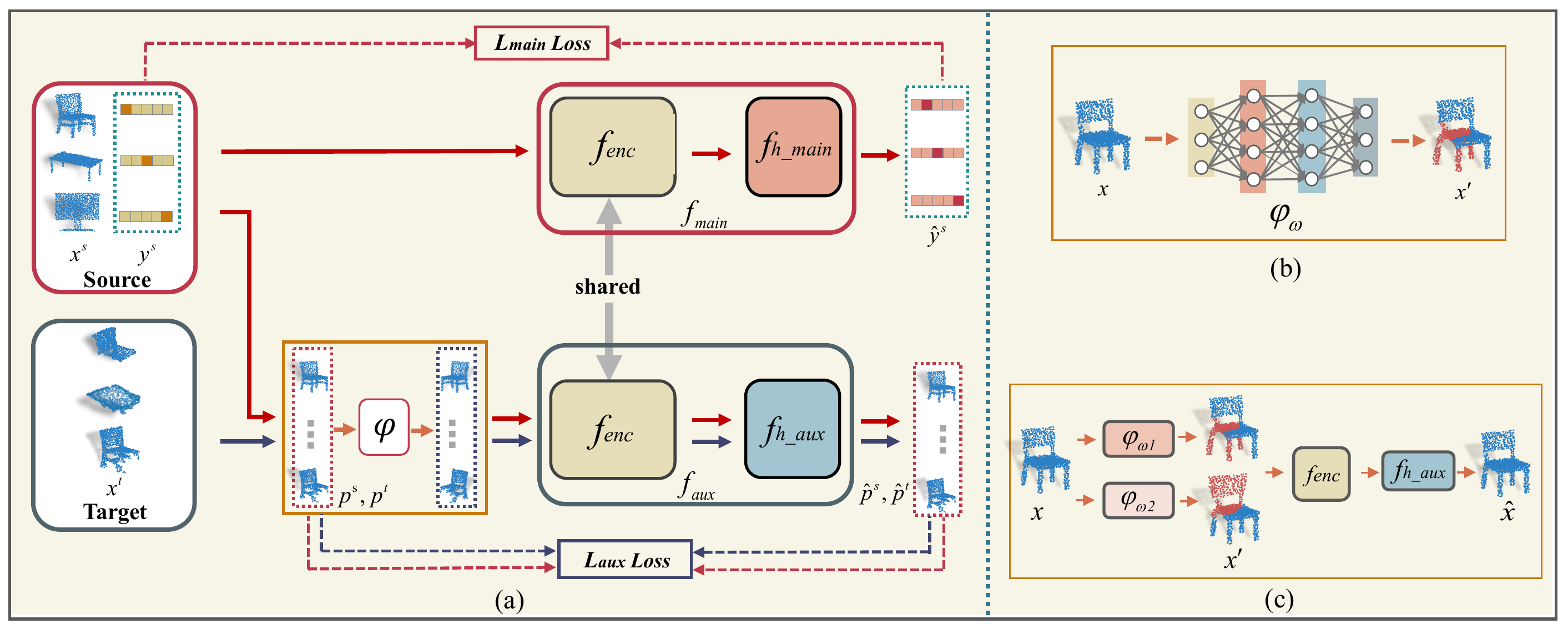}
   \caption{The proposed point cloud UDA framework. (a) The framework of the domain adaptation network based on self-supervised learning, where the red and blue arrows represent data flows from the source and target domains, respectively. (b) A learnable point cloud transformation network. (c) A self-supervised point cloud reconstruction task based on  multi-region transformation.}
   \label{figure2}
 \end{figure*}

\subsection{Self-Supervised Learning on Point Clouds}

Self-supervised learning utilizes pseudo labels generated from the data itself to train a learning model. It has long been a research focus in the field of machine learning and the readers are referred to \cite{53} for comprehensive review. Recently, there has been a series of work on point cloud self-supervised learning, most of which focus on proposing new self-supervised auxiliary tasks to improve performance on main tasks. There are several strategies commonly used to design  self-supervised tasks. The first strategy is to train a neural network to learn some features that can also be directly calculated from the point clouds. For example, \cite{37} introduced half-to-half point cloud prediction self-supervised task in a multi-angle scenario, where RNN was explored to predict the back half point cloud based on the front half. In \cite{38}, a novel geometric self-supervised learning task was proposed to predict point cloud local geometric information like normal and curvature. The second strategy is to first destroy a point cloud and then train a neural network to reconstruct it. Both \cite{13} and \cite{18} proposed self-supervised tasks agnostic to network architecture, in which a network is trained to reconstruct point clouds whose parts have been randomly rearranged. \cite{20} suggested simulating noise data and occlusion in real-world applications to construct the self-supervised reconstruction task, which is performed by replacing random selected regions of an original point cloud with new points sampled from a Gaussian distribution and then reconstructing the original point cloud. The third strategy is to design self-supervised tasks based on contrastive learning. For example, \cite{51} cut each point cloud into two parts and constructed contrastive learning by predicting whether the two parts belong to the same object. \cite{52} leveraged different modalities including image, point cloud, mesh and images from different views for the same 3D objects and they trained a model to differentiate positive pairs constructed from the same object and negative pairs constructed from different objects to make the model learn modal and view invariant features. There are other techniques to design self-supervised tasks. For example, \cite{19} transformed unstructured point clouds to sequences by space filling Morton-order curve and proposed a self-supervised  task in which a multi-layer RNN network was employed to predict the next point in the sequence. \cite{50} generated pseudo labels through clustering in feature space and combined self-supervised classification task with unsupervised clustering and reconstruction task to train a multi-scale encoder for point cloud and shape feature extraction. Nevertheless, in all these studies, the self-supervised tasks are hand-crafted, while in this paper, we propose the first learnable self-supervised task and show superior performance in point cloud UDA. 
\section{Method}

In this section, we first present the problem formulation of point cloud UDA and the UDA framework based on self-supervised task. Then, we explain the proposed point cloud destruction-reconstruction self-supervised task in detail, where point clouds are destroyed by a learnable nonlinear transformation. Finally, we propose a multi-region destruction strategy to extract more transferable domain-variant features. Please note that the problem formulation and the method are presented with point cloud classification as the main task, and the idea can be easily extended to point cloud segmentation. 
\subsection{Problem Formulation}

In UDA for point cloud classification, we have a source domain
$S=\{(x_i^s,y_i^s)\}_{i=1}^{n_s}$ with $n_s$ labeled point clouds and a target domain $T=\{{x_j^t}\}_{j=1}^{n_t}$ 
with $n_t $ unlabeled point clouds. Each point cloud $x_i^s $ has its  label $y_i^s$  in the source domain. 
The classes of the target domain are the same with those of the source domain, but the target domain sample 
$x_j^t $ is unlabeled. The objective is to train a classification network that generalizes well for classifying the target domain point clouds.

\subsection{UDA Framework Based on Self-Supervised Auxiliary Task }

\begin{figure*}[t]
   \centering
   \includegraphics[scale=0.85]{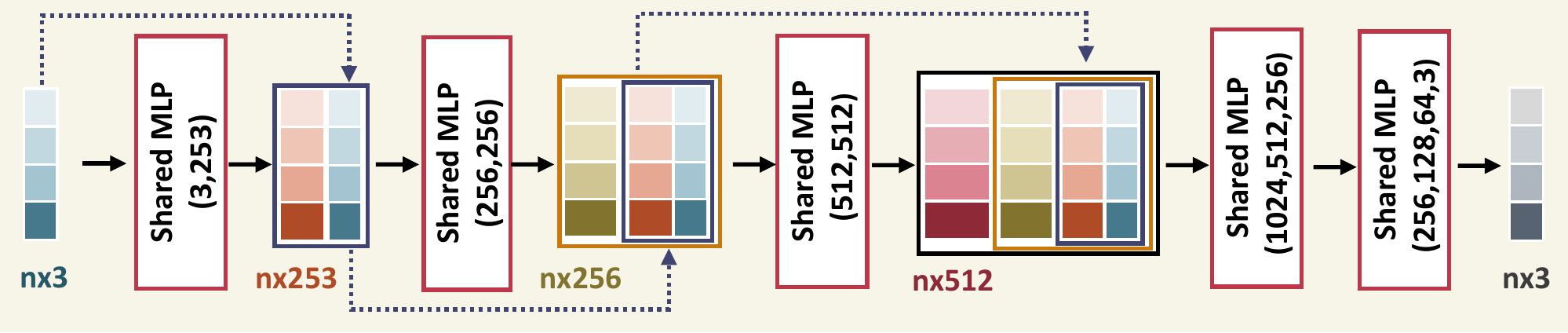}
   \caption{Point cloud transformation network. The blocks with different colors represent point cloud features, the dashed lines represent skip connections, and the values in the shared MLP module represent the numbers of neurons. Note that the batch normalization layer is applied after each layer of  shared MLP, and ReLU is used as the activation function.}
   \label{figure3}
 \end{figure*}

The architecture  of utilizing self-supervised tasks for UDA in \cite{18}  \cite{20} can be summarized as the 
framework shown in Figure \ref{figure2}  (a), which is also used in this study. The framework consists of a network $f_{main}$ for the main task of point cloud classification and an auxiliary network $f_{aux} $ for the 
self-supervised auxiliary task, where $f_{main} $ and $f_{aux}$ have their own heads   $f_{h\_main} $ and 
$f_{h\_aux}$, respectively, but share the same encoder $f_{enc}$. The network $f_{main}$ is trained by the main 
task loss $L_{main} $, which is defined as the cross-entropy loss  between predicted label ${\hat{y}}^s$ 
and the  corresponding ground truth label $y^s $ on the source domain data. For the auxiliary network, both the  source domain data $x^s$  and the target domain data $x^t $ are utilized for self-supervised learning. First, the original point 
clouds $x^s $ and $x^t $ are processed by a function $\varphi$,  and their pseudo labels $p^s$ and $p^t $
for the auxiliary self-supervised task can be generated from the original data and $\varphi$. Then, 
the auxiliary network is 
trained in a supervised manner with inputs as  $\varphi(x^s)$ and $\varphi(x^t)$ and their corresponding labels $p^s$ and
$p^t$. The loss to train the auxiliary network is denoted as $L_{aux}$. Hence, the final optimization 
objective is:

\begin{small}
\begin{align}
   &\min \sum_{i=1}^{n_{s}} L_{main}(f_{main}(x_{i}^{s}), y_{i}^{s})+
   \sum_{i=1}^{n_{s}} L_{aux}(f_{aux}(\varphi(x_{i}^{s})),p_{i}^{s}) \notag \\
   &\qquad  \qquad \qquad \quad +\sum_{j=1}^{n_{t}} L_{aux}(f_{aux}(\varphi(x_{j}^{t})), p_{j}^{t}) 
  \end{align}
\end{small}

\noindent where the first term is the loss of the main task on the source domain data, and the second and the third terms are the loss of the self-supervised auxiliary task on the source domain data and the target domain data, respectively.

In the above architecture, the purpose of performing the auxiliary task is to learn transferable features  by the encoder for the target domain, and this is achieved by training the encoder to extract features from both the source domain and the target domain data. In addition, local features tend to be more transferable across different point cloud datasets than global features. For instance, the overall shape of an aircraft varies greatly across datasets, but the differences between wings are relatively small, so the designed self-supervised task should encourage the encoder to focus on local features \cite{49}. In \cite{20}, Achituve et al. proposed transforming a portion of the point clouds, forcing the encoder to extract local features from the remaining component of the point cloud to perform the point cloud reconstruction task. Similarly, the 3D puzzle problem constructed in \cite{18} also enables the encoder to pay attention to local features. Following this idea, we propose a learnable self-supervised task to make the encoder focus more on local regions than on the whole point cloud.
\subsection{Self-Supervised Task Using A Learnable Transformation}

Many self-supervised tasks on point clouds rely on hand-crafted transformations of the point clouds, and different transformations will lead to different features being learned by the encoder. Transformations explored in literature include adding noise \cite{20}, removing a part of the point cloud \cite{20}, shuffling the subblocks of a point cloud \cite{18}, and so on. However, hand-crafted transformations are limited and nonflexible, and this may restrict the effectiveness of the trained encoder. In this section, we propose a learnable transformation $\varphi$  based on deep neural networks, as shown in Figure \ref{figure2} (b).

We construct a transformation network $\varphi_\omega $ parameterized by $\omega $ to transform the original point cloud 
$x$ into $x^{\prime}$. The auxiliary network $f_{aux} $ is applied to reconstruct the original point 
cloud from $x^{\prime}$, 
and the reconstructed point cloud is denoted as $\hat{x}$:

\begin{equation}
   x^{\prime}=\varphi_{\omega}(x) \label{eq2}
\end{equation}
 
\begin{equation}
   \hat{x}=f_{h_{\_{aux}}}\left(f_{enc}\left(x^{\prime}\right)\right) \label{eq3}
\end{equation}

The objective of the transformation network $\varphi_\omega $ is to maximize the chamfer distance between 
the original point cloud and the transformed point cloud, and the objective of the auxiliary 
reconstruction network, which consists of the shared encoder $f_{enc} $ and the head $f_{h\_aux}$, is to 
minimize the chamfer distance between the reconstructed point cloud and the original point cloud. 
So, the optimization objective of the auxiliary self-supervised task in formula (3) becomes:

\begin{equation}
   \min \lambda_{1} C D(\hat{x}, x)-\lambda_{2} C D\left(x^{\prime}, x\right)
\end{equation}
where $\lambda_1 $and $\lambda_2 $ are hyper-parameters, and $CD $ represents the chamfer distance 
between two point clouds. Training the transformation network $\varphi_\omega $ and the 
reconstruction network  $f_{enc}  \circ f_{h\_aux}$
in such an end-to-end way would enable $\varphi_\omega $ to learn a nonlinear and continuous transformation. 
Meanwhile, the encoder would also learn more robust features to reconstruct the original point 
cloud from the transformed one, thereby improving the transferability of the learned features 
and domain adaptability of the encoder.

\begin{figure}[t]
   \centering
   \includegraphics[scale=0.8]{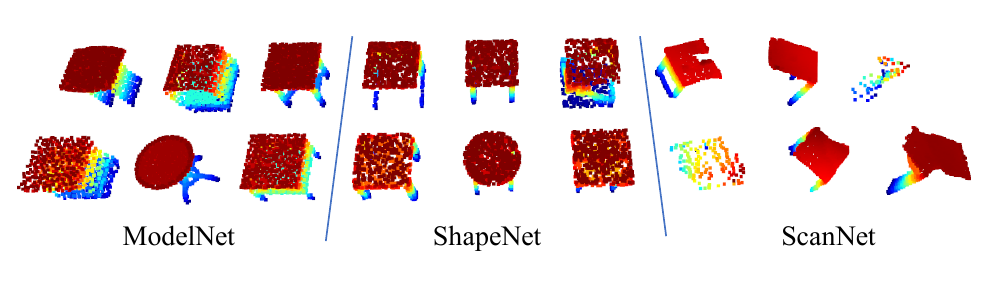}
   \caption{Examples of  point clouds in the  Point-DA dataset.}
   \label{figure}
 \end{figure}

 The transformation network is composed of multiple MLP layers with shared parameters as shown in Figure \ref{figure3}. Sharing parameters across all points ensures that neighboring points in the input point cloud get similar outputs after transformation so that the learned transformation is continuous and the global semantic information of the point cloud is maintained. Instead of directly outputting the coordinates of the points as in formula (4), we output the displacement of each point through the point cloud transformation network and control the scale of displacement by a  shift scale hyperparameter  $\alpha$ as follows:

\begin{equation}
   x^{\prime}=\alpha \cdot \varphi_{\omega}(x)+x
\end{equation}

\subsection{Multi-Region Transformation}
Experiments show that transforming the whole point cloud will make it very difficult to reconstruct the original point cloud, which may be due to the loss of most  local and contextual information during transformation.  To avoid this issue, we introduce a multi-region local point cloud transformation strategy based on our proposed learnable transformation network. 

As shown in Figure \ref{figure2} (c) , we employ a local transformation for the point cloud, so that the encoder 
can learn features from both the transformed and the untransformed regions. For one point cloud $x$, 
multiple transformation networks $\varphi_\omega$ with shared parameters are used to transform  
parts of the point cloud at different random locations to obtain several different transformed point 
clouds of the same object, and then the same reconstruction network $f_{enc}  \circ f_{h\_aux}$
is applied to reconstruct the original point cloud from the transformed ones. This strategy facilitates  the encoder to extract features from different parts of the objects for the reconstruction task.
\begin{table*}[t]
   \centering
   \caption{The classification accuracy (\%) on the PointDA-10 dataset with PointNet as the encoder.}
   \scalebox{0.95}{
   \begin{tabular}{p{7.25em}|ccccccc}
      \multicolumn{1}{c}{} &       &       &       &       &       &       &  \\
      \toprule
      \multirow{2}[2]{*}{\textbf{Method}} & \multicolumn{1}{c}{\textbf{ModelNet\ to}} & \multicolumn{1}{c}{\textbf{ModelNet to}} & \multicolumn{1}{c}{\textbf{ShapeNet to}} & \multicolumn{1}{c}{\textbf{ShapeNet to}} & \multicolumn{1}{c}{\textbf{ScanNet to}} & \multicolumn{1}{c}{\textbf{ScanNet to}} & \multicolumn{1}{c}{\multirow{2}[2]{*}{\textbf{Avg}}} \\
      \multicolumn{1}{c|}{} & \multicolumn{1}{c}{\textbf{ShapeNet}} & \multicolumn{1}{c}{\textbf{ScanNet}} & \multicolumn{1}{c}{\textbf{ModelNet}} & \multicolumn{1}{c}{\textbf{ScanNet}} & \multicolumn{1}{c}{\textbf{ModelNet}} & \multicolumn{1}{c}{\textbf{ShapeNet}} &  \\
      \midrule
      w/o Adapt & 80.2  & 43.1  & \textbf{75.8}  & 40.7  & 63.2  & 67.2    & 61.7 \\
      Rotate  & 81.6 & 48.2	& 64.6	& 49.0	& 48.0	& 63.0	& 59.1 \\
      PointDAN & 80.2  & 45.3  & 71.2  & 46.9  & 59.8  & 66.2  & 61.6 \\
      DefRec   & 80.0    & 46.0    & 68.5  & 41.7  & 63.0    & 68.2  & 61.2 \\
      DefRec+PCM  & 81.1  & 50.3  & 54.3  & 52.8  & 54.0    & \textbf{69.0}    & 60.3 \\
      Resort  & 81.6  & 49.7  & 73.6  & 41.9  & 65.9  & 68.1  & 63.5 \\
      \midrule
      Ours  & \textbf{82.5} & \textbf{52.7} & 73.8 & \textbf{53.8} & \textbf{67.4} & \textbf{69.0} & \textbf{66.5} \\
      \bottomrule
     \end{tabular}%
   }
   \label{tab1}%
  
 \end{table*}%

\begin{table*}[t]
   \centering
   \caption{The classification accuracy (\%) on the PointDA-10 dataset with DGCNN as the encoder.}
   \scalebox{0.95}{
      \begin{tabular}{p{7.25em}|ccccccc}
      \multicolumn{1}{c}{}   &       &       &       &       &       &       &  \\
     \toprule
     \multirow{2}[2]{*}{\textbf{Method}} & \multicolumn{1}{c}{\textbf{ModelNet\ to}} & \multicolumn{1}{c}{\textbf{ModelNet to}} & \multicolumn{1}{c}{\textbf{ShapeNet to}} & \multicolumn{1}{c}{\textbf{ShapeNet to}} & \multicolumn{1}{c}{\textbf{ScanNet to}} & \multicolumn{1}{c}{\textbf{ScanNet to}} & \multicolumn{1}{c}{\multirow{2}[2]{*}{\textbf{Avg}}} \\
     \multicolumn{1}{c|}{} & \multicolumn{1}{c}{\textbf{ShapeNet}} & \multicolumn{1}{c}{\textbf{ScanNet}} & \multicolumn{1}{c}{\textbf{ModelNet}} & \multicolumn{1}{c}{\textbf{ScanNet}} & \multicolumn{1}{c}{\textbf{ModelNet}} & \multicolumn{1}{c}{\textbf{ShapeNet}} &  \\
      \midrule
     \multicolumn{1}{p{7.25em}|}{w/o Adapt } & \multicolumn{1}{c}{81.7} & \multicolumn{1}{c}{42.9} & \multicolumn{1}{c}{72.2} & \multicolumn{1}{c}{44.2} & \multicolumn{1}{c}{67.3} & \multicolumn{1}{c}{65.1} & \multicolumn{1}{c}{62.2} \\
     \multicolumn{1}{p{7.25em}|}{Rotate } & \multicolumn{1}{c}{83.0} & \multicolumn{1}{c}{51.6} & \multicolumn{1}{c}{72.5} & \multicolumn{1}{c}{41.0} & \multicolumn{1}{c}{67.1} & \multicolumn{1}{c}{70.3} & \multicolumn{1}{c}{64.3} \\
     \multicolumn{1}{p{7.25em}|}{PointDAN } & \multicolumn{1}{c}{\textbf{83.9}} & \multicolumn{1}{c}{44.8} & \multicolumn{1}{c}{63.3} & \multicolumn{1}{c}{45.7} & \multicolumn{1}{c}{43.6} & \multicolumn{1}{c}{56.4} & \multicolumn{1}{c}{56.3} \\
     \multicolumn{1}{p{7.25em}|}{RS } & \multicolumn{1}{c}{81.5} & \multicolumn{1}{c}{35.2} & \multicolumn{1}{c}{71.9} & \multicolumn{1}{c}{39.8} & \multicolumn{1}{c}{61.0} & \multicolumn{1}{c}{63.6} & \multicolumn{1}{c}{58.8} \\
     \multicolumn{1}{p{7.25em}|}{DefRec } & \multicolumn{1}{c}{83.3} & \multicolumn{1}{c}{46.6} & \multicolumn{1}{c}{79.8} & \multicolumn{1}{c}{49.9} & \multicolumn{1}{c}{70.7} & \multicolumn{1}{c}{64.4} & \multicolumn{1}{c}{65.8} \\
     \multicolumn{1}{p{7.25em}|}{DefRec+PCM } & \multicolumn{1}{c}{81.7} & \multicolumn{1}{c}{51.8} & \multicolumn{1}{c}{78.6} & \multicolumn{1}{c}{54.5} & \multicolumn{1}{c}{\textbf{73.7}} & \multicolumn{1}{c}{71.1} & \multicolumn{1}{c}{68.6} \\
     \midrule
     \multicolumn{1}{p{7.25em}|}{Ours} & \multicolumn{1}{c}{82.8} & \multicolumn{1}{c}{\textbf{56.3}} & \multicolumn{1}{c}{\textbf{81.7}} & \multicolumn{1}{c}{\textbf{54.8}} & \multicolumn{1}{c}{72.9} & \multicolumn{1}{c}{\textbf{71.7}} & \multicolumn{1}{c}{\textbf{70.0}} \\
     \bottomrule
     \end{tabular}%
     }
   \label{tab2}%
 \end{table*}%

\section{Experiment}
\subsection{ Point Cloud Classification UDA}

\noindent \textbf{Dataset}.  We evaluated our method on the PointDA-10 dataset \cite{49} specifically designed for point cloud DA; this dataset contains 10 shared classes from three widely used point cloud datasets: ModelNet \cite{72}, ShapeNet \cite{73} and ScanNet \cite{74}, as shown in Figure \ref{figure}. The point clouds in ModelNet and ShapeNet were sampled from 3D CAD models, while point clouds in ScanNet were sampled from scanned and reconstructed real-world indoor scenes. As in \cite{20}, we rotate the ScanNet and ShapeNet models $90$ around their $Y$ axis, so that in all three datasets, the upward direction of each model corresponds to the positive direction of $Z$ axis.  There are obvious domain gaps between the three datasets, as shown in Figure \ref{figure}. The point clouds in both ModelNet and ShapeNet are complete surface points of 3D objects, but the objects in the two sets have different shapes and styles, while the point clouds in ScanNet are partial surface points with noise. UDA experiments were performed by choosing one of them as the source domain and another as the target domain, and this resulted in six UDA scenarios, where the performance was measured in terms of classification accuracy on the target domain. We split the official training set with 80\% for training and 20\% for validation, and used the official test sets for testing.

\noindent \textbf{Implementation Details}. Our model was trained on a single NVIDIA V100 GPU based on the deep learning library PyTorch \cite{70}. The ADAM \cite{71} optimizer with a learning rate of 0.001 and a weight decay of 0.0005 was applied under a cosine annealing learning rate scheduler. Early-stop mechanism was applied on validation dataset to avoid over-fitting. The batch size and number of epochs were set as 16 and 150, respectively. During training, we applied rotations about the $Z $ axis and random jittering with standard deviation and clip parameters of 0.01 and 0.02, respectively, for data augmentation. The model with the highest classification accuracy on the source domain validation dataset was preserved to evaluate the performance of that model on the target domain test dataset.

\noindent \textbf{Network Options}. Any point cloud encoding network can be used as the encoder in our method, 
and we experimented on two networks: PointNet and DGCNN. The main task head network and the point cloud 
reconstruction head network are both fully connected networks. In the multi-region point cloud transformation 
strategy, we constructed two transformation networks with shared parameters to transform the 512 nearest 
neighbors of a randomly selected point with a shift scale $ \alpha$ of 0.05. Since the point cloud transformation 
network is easier to train than the reconstruction network, we set $\lambda_1 $ and $\lambda_2 $ as 1 and 10, respectively.

\noindent \textbf{Random Cropping}. ModelNet and ShapeNet contain point clouds of complete surfaces of objects, while ScanNet contains scanned and reconstructed real-world point clouds, which are often incomplete surfaces of objects. Due to the shape differences between them, we applied a random cropping strategy as in \cite{69} for data augmentation when ScanNet was chosen as the target domain, where the point clouds of the source domain were randomly cropped by a plane with a random direction, and 70\% of the point cloud was retained.

\noindent \textbf{Classification Results}. We compared our proposed method with other competitive point cloud UDA methods, including PointDAN \cite{49}, DefRec \cite{20}, Resort \cite{18} and RS \cite{13}. In addition, we compared with another two baselines methods. The first baseline is to train the main task network with the source domain data and then directly use it on the target domain without adaptation. The second baseline utilizes the same architecture as our proposed method but uses rotation prediction as the self-supervised task, and it is denoted as Rotate. In addition, these UDA methods are encoder-independent, so we compare the results with either PointNet or DGCNN as the encoder to evaluate the generality of the comparing UDA methods. The results with PointNet and DGCNN as the encoder are shown in Table \ref{tab1} and Table \ref{tab2}, respectively. We reproduced PointDAN's results. Because the code of Resort \cite{18} is not publicly available, we only use the results reported in the original paper with PointNet as the encoder. The results of DefRec and RS are from \cite{20}, and PCM refers to an extension with data augmentation adopted in \cite{20}.

Our proposed method achieves the highest average classification accuracy with both encoders. In Table \ref{tab1} with PointNet as the encoder, our method achieves the highest accuracy on five out of six UDA scenarios, and w/o Adapt achieves the highest accuracy on another scenario. In Table \ref{tab2} with DGCNN as the encoder, our method achieves the highest classification accuracy on four out of six UDA scenarios, while DefRec + PCM achieves the highest accuracy on one scenario and PointDAN achieves the highest accuracy on one scenario. With PointNet and DGCNN as the encoder, the average accuracy of our method is 4.8\% and 7.8\% higher than the baseline without adaptation, which indicates the efficiency of the proposed adaptation method. By comparing the results of the proposed method and Rotate, we can see that the proposed destruction-reconstruction self-supervised task is more effective than the rotation prediction task.  By comparing the results of Table \ref{tab1} and Table \ref{tab2}, we can find that our proposed method is more agnostic to different encoders.  
\begin{table*}[t]
   \centering
   \caption{Ablation study}
   \scalebox{0.92}{
     \begin{tabular}{ccccrrrrrrr}
           &       &       &       &       &       &       &       &       &       &  \\
     \midrule
     \multicolumn{1}{c|}{\multirow{2}[2]{*}{\textbf{SSL}}} & \multicolumn{1}{c|}{\multirow{2}[2]{*}{\textbf{Multi}}} & \multicolumn{1}{c|}{\multirow{2}[2]{*}{\textbf{Crop}}} & \multicolumn{1}{c|}{\multirow{2}[2]{*}{\textbf{DGCNN}}} & \multicolumn{1}{c}{\textbf{ModelNet\ to}} & \multicolumn{1}{c}{\textbf{ModelNet to}} & \multicolumn{1}{c}{\textbf{ShapeNet to}} & \multicolumn{1}{c}{\textbf{ShapeNet to}} & \multicolumn{1}{c}{\textbf{ScanNet to}} & \multicolumn{1}{c}{\textbf{ScanNet to}} & \multicolumn{1}{c}{\multirow{2}[2]{*}{\textbf{Avg}}} \\
     \multicolumn{1}{c|}{} & \multicolumn{1}{c|}{} & \multicolumn{1}{c|}{} & \multicolumn{1}{c|}{} & \multicolumn{1}{c}{\textbf{ShapeNet}} & \multicolumn{1}{c}{\textbf{ScanNet}} & \multicolumn{1}{c}{\textbf{ModelNet}} & \multicolumn{1}{c}{\textbf{ScanNet}} & \multicolumn{1}{c}{\textbf{ModelNet}} & \multicolumn{1}{c}{\textbf{ShapeNet}} &  \\
     \midrule
     \multicolumn{1}{c|}{} & \multicolumn{1}{c|}{} & \multicolumn{1}{c|}{} & \multicolumn{1}{c|}{} & \multicolumn{1}{c}{80.2} & \multicolumn{1}{c}{43.1} & \multicolumn{1}{c}{75.8} & \multicolumn{1}{c}{40.7} & \multicolumn{1}{c}{63.2} & \multicolumn{1}{c}{67.2} & \multicolumn{1}{c}{61.7} \\
     \midrule
     \multicolumn{1}{c|}{$\surd$} & \multicolumn{1}{c|}{} & \multicolumn{1}{c|}{} & \multicolumn{1}{c|}{} & \multicolumn{1}{c}{82.6} & \multicolumn{1}{c}{48.8} & \multicolumn{1}{c}{74.0} & \multicolumn{1}{c}{44.7} & \multicolumn{1}{c}{63.1} & \multicolumn{1}{c}{67.9} & \multicolumn{1}{c}{63.5} \\
     \multicolumn{1}{c|}{$\surd$} & \multicolumn{1}{c|}{$\surd$} & \multicolumn{1}{c|}{} & \multicolumn{1}{c|}{} & \multicolumn{1}{c}{82.5} & \multicolumn{1}{c}{47.8} & \multicolumn{1}{c}{73.8} & \multicolumn{1}{c}{46.2} & \multicolumn{1}{c}{67.4} & \multicolumn{1}{c}{69.0} & \multicolumn{1}{c}{64.5} \\
     \multicolumn{1}{c|}{$\surd$} & \multicolumn{1}{c|}{$\surd$} & \multicolumn{1}{c|}{$\surd$} & \multicolumn{1}{c|}{} & \multicolumn{1}{c}{82.5} & \multicolumn{1}{c}{52.7} & \multicolumn{1}{c}{73.8} & \multicolumn{1}{c}{53.8} & \multicolumn{1}{c}{67.4} & \multicolumn{1}{c}{69.0} & \multicolumn{1}{c}{66.5} \\
     \multicolumn{1}{c|}{$\surd$} & \multicolumn{1}{c|}{$\surd$} & \multicolumn{1}{c|}{} & \multicolumn{1}{c|}{$\surd$} & \multicolumn{1}{c}{\textbf{82.8}} & \multicolumn{1}{c}{46.7} & \multicolumn{1}{c}{\textbf{81.7}} & \multicolumn{1}{c}{51.3} & \multicolumn{1}{c}{\textbf{72.9}} & \multicolumn{1}{c}{\textbf{71.7}} & \multicolumn{1}{c}{67.9} \\
     \multicolumn{1}{c|}{$\surd$} & \multicolumn{1}{c|}{} & \multicolumn{1}{c|}{$\surd$} & \multicolumn{1}{c|}{$\surd$} & \multicolumn{1}{c}{80.6} & \multicolumn{1}{c}{52.2} & \multicolumn{1}{c}{72.1} & \multicolumn{1}{c}{50.5} & \multicolumn{1}{c}{63.1} & \multicolumn{1}{c}{68.4} & \multicolumn{1}{c}{64.5} \\
     \multicolumn{1}{c|}{$\surd$} & \multicolumn{1}{c|}{$\surd$} & \multicolumn{1}{c|}{$\surd$} & \multicolumn{1}{c|}{$\surd$} & \multicolumn{1}{c}{\textbf{82.8}} & \multicolumn{1}{c}{\textbf{56.3}} & \multicolumn{1}{c}{\textbf{81.7}} & \multicolumn{1}{c}{\textbf{54.8}} & \multicolumn{1}{c}{\textbf{72.9}} & \multicolumn{1}{c}{\textbf{71.7}} & \multicolumn{1}{c}{\textbf{70.0}} \\
      \bottomrule
        &       &       &       &       &       &       &       &       &       &  \\
     \end{tabular}%
   }
   \label{tab3}%
 \end{table*}%

\subsection{ Point Cloud Segmentation UDA}

\noindent \textbf{Dataset and Implementation}. We then evaluated our method on PointSegDA dataset proposed in \cite{20} for point cloud segmentation UDA. The dataset contains human body point clouds collected from four datasets: ADOBE, FAUST, MIT and SCAPE. The point clouds in these datasets have different body poses and shapes, but they are all segmented into the same eight parts. UDA experiments were performed by choosing one of them as the source domain and another as the target domain, which resulted in 12 UDA scenarios. The performance was measured in terms of mean Intersection over Union (IoU) on the target domain. We split the training dataset with 80\% for training and 20\% for validation, and used the official test sets for testing. DGCNN was used as the encoder in this experiment and the segmentation head  was implemented using four 1D convolutional layers with size [256, 256 ,128, 8].

\noindent \textbf{Segmentation Results}. We compare our method with the baseline without adaptation, RS\cite{13} and DefRec \cite{20}. The results are shown in Table \ref{tab4}. Our method achieved the highest average IoU and the highest IoUs in eight out of 12 UDA scenarios. Compared with the baseline without adaptation, we can observe that significant improvement is achieved in most scenarios after employing our domain adaption method.

\begin{table*}[htbp]
   \centering
   \caption{The mean IoU on PointSegDA dataset with DGCNN as the encoder.}
   \scalebox{0.95}{
     \begin{tabular}{p{5.5em}|ccccccccccccc}
     \toprule
     \multirow{2}[2]{*}{\textbf{Method}} & \multicolumn{1}{p{2.75em}}{\textbf{FAUST  to   }} & \multicolumn{1}{p{2.565em}}{\textbf{FAUST to }} & \multicolumn{1}{p{2.565em}}{\textbf{FAUST to }} & \multicolumn{1}{p{2.375em}}{\textbf{MIT    to   }} & \multicolumn{1}{p{2.5em}}{\textbf{MIT     to }} & \multicolumn{1}{p{2.5em}}{\textbf{MIT      to   }} & \multicolumn{1}{p{2.69em}}{\textbf{ADOBE to }} & \multicolumn{1}{p{2.625em}}{\textbf{ADOBE to }} & \multicolumn{1}{p{2.44em}}{\textbf{ADOBE to }} & \multicolumn{1}{p{2.565em}}{\textbf{SCAPE to }} & \multicolumn{1}{p{2.375em}}{\textbf{SCAPE to   }} & \multicolumn{1}{p{2.44em}}{\textbf{SCAPE to }} & \multicolumn{1}{c}{\multirow{2}[2]{*}{\textbf{Avg}}} \\
     \multicolumn{1}{l|}{} & \multicolumn{1}{p{2.75em}}{\textbf{MIT}} & \multicolumn{1}{p{2.565em}}{\textbf{ADOBE}} & \multicolumn{1}{p{2.565em}}{\textbf{SCAPE}} & \multicolumn{1}{p{2.375em}}{\textbf{FAUST}} & \multicolumn{1}{p{2.5em}}{\textbf{ADOBE}} & \multicolumn{1}{p{2.5em}}{\textbf{SCAPE}} & \multicolumn{1}{p{2.69em}}{\textbf{FAUST}} & \multicolumn{1}{p{2.625em}}{\textbf{MIT}} & \multicolumn{1}{p{2.44em}}{\textbf{SCAPE}} & \multicolumn{1}{p{2.565em}}{\textbf{FAUST}} & \multicolumn{1}{p{2.375em}}{\textbf{MIT}} & \multicolumn{1}{p{2.44em}}{\textbf{ADOBE}} &  \\
     \midrule
     w/o Adapt & 60.9  & 78.5  & 66.5  & 33.6  & 26.6  & 69.9  & 38.5  & 31.2  & 30.0    & 64.5  & \textbf{74.1} & 68.4  & 53.6 \\
     RS    & 60.7  & 78.7  & 66.9  & 38.4  & 59.6  & 70.4  & 44.0    & 30.4  & 36.6  & 65.3  & 70.7  & \textbf{73.0} & 57.9 \\
     DefRec & \textbf{61.8} & 79.7  & 67.4  & 40.1  & \textbf{67.1} & \textbf{72.6} & 42.5  & 28.9  & 32.2  & 66.2  & 66.4  & 72.2  & 58.1 \\
     DefRec+PCM & 60.9  & 78.8  & 63.6  & 48.6  & 48.1  & 70.1  & 46.9  & 33.2  & 37.6  & 62.6  & 66.3  & 66.5  & 56.9 \\
     Ours  & \textbf{61.8} & \textbf{80.3} & \textbf{68.5} & \textbf{56.6} & 60.8  & 67.8  & \textbf{52.3} & \textbf{38.6} & \textbf{41} & \textbf{66.6} & 67.4  & 68.0    & \textbf{60.8} \\
     \bottomrule
     \end{tabular}%
   }
   \label{tab4}%
 \end{table*}%

\subsection{Ablation Study}

To verify the effects of different modules proposed in our method, we conduct a detailed ablation study, and the quantitative results are summarized in Table \ref{tab3}. SSL represents applying a self-supervised task to train the network, Multi represents applying the multi-region transformation strategy, Crop represents applying the random cropping data augmentation process when the target domain is ScanNet, and DGCNN represents using DGCNN instead of PointNet as the encoder. The first row shows the results obtained by the baseline model without adaptation.

From Table \ref{tab3}, we can see that even when applying the proposed self-supervised task alone, our average classification accuracy can reach 63.5\%, which is higher than those of the baseline and PointDAN  \cite{49}. In this situation, PointNet is the encoder, and the classification accuracy achieved by using the proposed self-supervised network is higher than that obtained using the deformation-based self-supervised task in DefRec \cite{20} and on par with the results in Resort \cite{18}. This result shows that our self-supervised task can indeed handle the domain distribution alignment more elaborately than the other methods, thus achieving better domain adaptation. After adopting the multi-region transformation strategy, the accuracy is further improved to 64.5\%. In the case with ScanNet as the target domain, the random cropping augmentation for the source domain data reduces the domain gap between the source domain and the target domain, thus improving the accuracy of ModelNet to ScanNet and ShapeNet to ScanNet by 4.9\% and 7.6\%, respectively. In this way, the obtained average accuracy (66.5\%) becomes the highest among those of all competitive methods with PointNet as the encoder. Finally, when DGCNN is applied as the encoder, our method achieves the state-of-the-art average accuracy of 70.0\%.

In addition, we also experimented on using Crop for the DefRec+PCM method, but its performance changes little. Concretely, the classification accuracy for the ModelNet to ScanNet decreased from 55.5\% to 54.4\%, and the classification accuracy for the ShapeNet to ScanNet increased from 53.4\% to 53.7\%.

\subsection{ Learned Transformation as Data Augmentation}

In this work, we propose a learnable self-supervion task, in which a transformation is learned to non-linearly transform a part of a point cloud. Actually, besides being used in the domain adaptation framework, the learned transformation can also be used as a data augmentation strategy. In this section, we conduct the following experiment to study if it is an effective data augmentation strategy and how is its performance when compared to the domain adaptation strategy. In this experiment, PointNet is used as the encoder. 

For each of the six domain adaptation scenarios, we do the following experiment. First, we learn a non-linear transformation by using the destruction-reconstruction self-supervised task on on the source domain data. Then, we train a classification network on the source domain data, in which the learned transformation is used as data augmentation. Finally, the learned classification network is used directly on classify the target domain point cloud. We also experiment on using the transformation in \cite{18} as augmentation. The results are shown in Table \ref{tab5}. In most scenarios, our data augmentation method achieves equal or higher accuracy than \cite{18} or without augmentation. This indicates that the transformation does improve the generalization ability of the trained model when it is used as data augmentation. However, when comparing with the results in Table \ref{tab1}, we can see that the domain adaptation strategy is better than the data augmentation strategy (Avg accuracy 66.5\% vs. 63.0\%). 

\begin{table}[htbp]
   \centering
   \caption{The classification accuracy (\%) on the PointDA-10 dataset}
   \scalebox{0.6}{
     \begin{tabular}{p{4.4em}|ccccccc}
     \toprule
     \multirow{2}[2]{*}{\textbf{Method}} & \multicolumn{1}{p{4.315em}}{\textbf{ModelNet to }} & \multicolumn{1}{p{4.19em}}{\textbf{ModelNet to}} & \multicolumn{1}{p{4.19em}}{\textbf{ShapeNet to}} & \multicolumn{1}{p{4.19em}}{\textbf{ShapeNet to}} & \multicolumn{1}{p{4.19em}}{\textbf{ScanNet to}} & \multicolumn{1}{p{4.19em}}{\textbf{ScanNet to}} & \multicolumn{1}{c}{\multirow{2}[2]{*}{\textbf{Avg}}} \\
     \multicolumn{1}{l|}{} & \multicolumn{1}{p{4.315em}}{\textbf{ShapeNet}} & \multicolumn{1}{p{4.19em}}{\textbf{ScanNet}} & \multicolumn{1}{p{4.19em}}{\textbf{ModelNet}} & \multicolumn{1}{p{4.19em}}{\textbf{ScanNet}} & \multicolumn{1}{p{4.19em}}{\textbf{ModelNet}} & \multicolumn{1}{p{4.19em}}{\textbf{ShapeNet}} &  \\
     \midrule
     w/o Adapt & 80.2  & 43.1  & \textbf{75.8} & 40.7  & 63.2  & 67.2  & 61.7 \\
     DefRec & 81.0  & \textbf{44.9} & 73.0  & 42.2  & 61.7  & \textbf{68.8} & 61.9 \\
     Ours  & \textbf{81.2} & 43.1  & 75.1  & \textbf{43.2} & \textbf{66.7} & \textbf{68.8} & \textbf{63.0 } \\
     \bottomrule
     \end{tabular}%
   }
   \label{tab5}%
 \end{table}%
 
 
\section{Conclusion}
To address the UDA problem for point clouds, we propose a novel learnable self-supervised task that helps the adapted neural network extract transferable features. Specifically, we propose a learnable point cloud transformation and use it in a point cloud destruction-reconstruction self-supervised auxiliary task, and we apply it in a UDA framework with a multitask learning architecture. We train the main task network and the auxiliary task network, which share an encoder, so that the encoder extracts features that are highly transferable to the target domain. We further propose a multi-region transformation strategy to make the network focus on local features, which are more transferable. New state-of-the-art performance is achieved on the point cloud classification UDA benchmark PointDA-10 and point cloud segmentation UDA benchmark PointSegDA. We think that the proposed learnable self-supervised task can be applied in other self-supervised learning and semi-supervised learning studies. 

{\small
\bibliographystyle{ieee_fullname}
\bibliography{ref}
}

\end{document}